# FLECS: Planning with a Flexible Commitment Strategy

**Manuela Veloso** VELOSO@CS.CMU.EDU
**Peter Stone** PSTONE@CS.CMU.EDU
*Department of Computer Science, Carnegie Mellon University*
*Pittsburgh, PA 15213-3891 USA*

## Abstract

There has been evidence that least-commitment planners can efficiently handle planning problems that involve difficult goal interactions. This evidence has led to the common belief that delayed-commitment is the "best" possible planning strategy. However, we recently found evidence that eager-commitment planners can handle a variety of planning problems more efficiently, in particular those with difficult operator choices. Resigned to the futility of trying to find a universally successful planning strategy, we devised a planner that can be used to study which domains and problems are best for which planning strategies. In this article we introduce this new planning algorithm, FLECS, which uses a **FLE**xible **C**ommitment **S**trategy with respect to plan-step orderings. It is able to use any strategy from delayed-commitment to eager-commitment. The combination of delayed and eager operator-ordering commitments allows FLECS to take advantage of the benefits of explicitly using a simulated execution state and reasoning about planning constraints. FLECS can vary its commitment strategy across different problems and domains, and also during the course of a single planning problem. FLECS represents a novel contribution to planning in that it explicitly provides the choice of which commitment strategy to use while planning. FLECS provides a framework to investigate the mapping from planning domains and problems to efficient planning strategies.

## 1. Introduction

General-purpose planning has a long history of research in Artificial Intelligence. Several different planning algorithms have been developed ranging from the pioneering GPS (Ernst & Newell, 1969) to a variety of recent algorithms in the SNLP (McAllester & Rosenblitt, 1991) family. At the most basic level, the purpose of planning is to find a sequence of actions that change an *initial state* into a state that satisfies a *goal statement*. Planners use the actions provided in their domain representations to try to achieve the goal. However different planners use different means to this end.

Faced with a variety of different planning algorithms, some planning researchers, including these authors, have been increasingly curious to compare different planning methodologies. Although general-purpose planning is known to be undecidable (Chapman, 1987), it has been a common belief that least-commitment planning is the "best," i.e., the most efficient planning strategy for most planning problems. This belief is based on evidence that least-commitment planners can efficiently handle planning problems that involve difficult plan step interactions (Barrett & Weld, 1994; Kambhampati, 1994; Minton, Bresina, & Drummond, 1991). Delayed commitments, in particular to step orderings, allow the plan steps to remain unordered until the interactions are visible.[1] In similar situations, eager-commitment planners may encounter severe efficiency problems with early commitments to incorrect orderings.

Recently we engaged in an investigation of other sorts of planning problems which would be handled efficiently by other planning strategies. Since all planning is driven by heuristics, we identified different sets of heuristics that correspond to different planning methods. We designed sets of planning domains and problems to test different planning strategies. While studying the impact of these different strategies in different kinds of planning problems, we came across evidence that eager-commitment planners can efficiently handle a variety of planning problems, in particular those with difficult operator choices (Stone, Veloso, & Blythe, 1994). The up-to-date state allows them to make informed planning choices, particularly in terms of the operator alternatives available. In similar situations, delayed-commitment planners may need to backtrack over incorrect operator choices (Veloso & Blythe, 1994). We came to believe that no planner was consistently better than all others across different domains and problems.

Resigned to the futility of trying to find a universally successful planning strategy, we felt the need to study which domains and problems were best suited to which planning methods.[2] In order to do so, we devised and implemented a planner that can use any operator-ordering commitment strategy along the continuum between, on the one extreme delayed commitment, and on the other, eager commitment. This planner is completely flexible along one dimension of planning heuristics: operator-ordering commitments. Our main contribution in this paper is to completely describe this planning algorithm and to put it forth as a tool for studying the mapping between heuristics and domains or problems. Rather than risking the possibility that the planner itself might get overlooked if it were relegated to an "architecture" section of a future paper, we present FLECS and its underlying philosophy as a contribution in its own right.

The continuum of heuristics that can be explored by our planning algorithm lies between the operator-ordering commitment strategies of delayed-commitment and eager-commitment backward-chaining planners, which we now situate within a broad range of planning and problem solving methods. One possible planning strategy is to search all the possible states that can be reached from the initial state to find one that satisfies the goal. This method, called *progression* or *forward-chaining*, can be very impractical. There are often too many accessible states in the world to efficiently search the complete state space. As an alternative, several planners constrain their search by using *regression*, or *backward-chaining*. Rather than considering all possible actions that could be executed in the initial state and searching recursively forward through the state space, they search backwards from the goal. Their search is driven by the set of actions that can directly achieve the goal.

There are two main ways of performing backward-chaining. Several planners do regression by searching the space of possible plans. Planners, such as NOAH, TWEAK, SNLP,

1. Least-commitment planners really *delay* commitments to plan step orderings and to variable bindings. Throughout this article we use the term *delayed* commitment to contrast with eager commitment in the context of step orderings.
2. Similar concerns regarding different constraint satisfaction algorithms have led recently to the design of the Multi-Tac architecture (Minton, 1993). This system investigates a given problem to find a combination of heuristics from a collection of available ones to solve the problem in an efficient way.

and their descendants (Chapman, 1987; McAllester & Rosenblitt, 1991; McDermott, 1978; Sacerdoti, 1977; Tate, 1977; Wilkins, 1984) are plan-space planners that use a delayed-commitment strategy. In particular, they delay the decision of ordering operators as long as possible. Consequently, the planner reasons from the initial state and from a set of constraints that are regressed from the goal. On the other hand, planners such as gps, strips, and the prodigy family (Carbonell, Knoblock, & Minton, 1990; Fikes & Nilsson, 1971; Rosenbloom, Newell, & Laird, 1990) use an eager-commitment strategy.[3] They use backward-chaining to select plan steps relevant to the goals. These eager-commitment planners make explicit use of an internal representation of the state of the world (their *internal state*) and order operators when possible so that they can reason from an updated version of this state. They trade the risk of eager commitment for the benefits of using an explicit updated planning state.

In this article we introduce a planning algorithm, flecs, that uses a **FLE**xible **C**ommitment **S**trategy with respect to operator orderings. flecs is designed to provide us and other planning researchers with a framework to investigate the mapping from domains and problems to efficient planning strategies. This algorithm represents a novel contribution to planning in that it introduces explicitly the choice of the commitment strategy. This ability to change its commitment strategy makes it useful for studying the tradeoffs between delayed and eager commitments. flecs is a descendant of prodigy4.0 and its current implementation is directly on top of prodigy4.0. It extends prodigy4.0 by reasoning explicitly about ordering alternatives and by having the ability to change its commitment strategy across different problems and domains, and also during the course of a single planning problem.[4]

This article gradually introduces flecs. Section 2 gives a top-level view of the algorithm and describes the different ways in which flecs makes use of a uniquely specified state of the world. Section 3 introduces the concepts used by the flecs algorithm. We provide an annotated example to illustrate the details of the planning concepts defined. Section 4 presents flecs's planning algorithm in full detail and explains the algorithm step by step. We discuss different heuristics to guide flecs's choices, in particular the flexible choice of commitment strategy. We analyze the advantages and disadvantages of delayed and eager plan step ordering commitments. Section 5 shows specific examples of planning domains and problems that we devised, which support the need for the use of flecs's flexible commitment strategy. We performed an empirical analysis on planning performance in these domains. The corresponding empirical results demonstrate the tradeoffs discussed and show evidence that flexible commitment is necessary. Finally Section 6 draws conclusions from this work.

3. Planners in the prodigy family include prodigy2.0 (Minton, Knoblock, Kuokka, Gil, Joseph, & Carbonell, 1989), NoLimit (Veloso, 1989), and prodigy4.0 (Carbonell, Blythe, Etzioni, Gil, Joseph, Kahn, Knoblock, Minton, Pérez, Reilly, Veloso, & Wang, 1992). NoLimit and prodigy4.0, as opposed to prodigy2.0, do not require the linearity assumption of goal independence and their search spaces are complete (Fink & Veloso, 1994). They also have some control over their commitment choices as opposed to the other earlier total-order planners.
4. We found that we needed a new name for our algorithm as flecs represents a significant change in philosophy and implementation from prodigy4.0.

## 2. A Top-Level View of FLECS

PRODIGY4.0 and FLECS differ most significantly from other state-of-the-art planning systems in that they search for a solution to a planning problem by combining backward-chaining (or regression) and simulation of plan execution (Fink & Veloso, 1994). While back-chaining, they can commit to a total-ordering of plan steps so as to make use of a uniquely specified world state. These planners maintain an internal representation of the state and update it by simulating the execution of operators found relevant to the goal by backward-chaining. Note that *simulating* execution while planning differs from interleaving planning and execution, since the option of "un-simulating," or rolling back, must remain open. Interleaved planning and execution is generally done by separate modules for planning, monitoring, executing, and replanning (Ambros-Ingerson & Steel, 1988). FLECS can either delay or eagerly carry out the plan simulation. In this way, our planning algorithm has the flexibility of both being able to delay operator-ordering commitments *and* being able to use the effects of previously selected operators to help determine which goals to plan for next and which operators to use to achieve these goals. In short, it can emulate both delayed-commitment planners and eager-commitment planners.

Table 1 shows the top-level view of the FLECS algorithm.

---

1. Initialize.
2. Terminate if the goal statement has been satisfied.
3. Compute the pending goals and applicable operators.
   - Pending goals are the yet-to-be-achieved preconditions of operators that have been selected to be in the plan.
   - Applicable operators are those that have all their preconditions satisfied in the current state.
5. Choose to subgoal or apply: *(backtrack point)*
   - To subgoal, go to step 6.
   - To apply, go to step 7.
6. Select a pending goal *(no backtrack point)* and an operator that can achieve it *(backtrack point)*; go to step 3.
7. Change the state as specified by an applicable operator *(backtrack point)*; go to step 2.

---

Table 1: A top-level view of FLECS. The step numbers here are made to correspond with the step numbers in the detailed version of the algorithm presented in Table 2 (Section 4), which refines these steps and adds an additional necessary step 4.

All the terms used in this table are fully described along with the detailed version of the algorithm in Section 4. In this section we focus on two main characteristics of this algorithm, namely its use of an internal state and its flexibility with respect to commitment strategies.

## 2.1 The Use of a Simulated Planning State

FLECS uses its internal state for at least four purposes. First, it terminates when every goal from the given problem is satisfied in the current version of the state (the *current state*): at this point, a complete plan (the sequence of operators that transformed the initial state into the current state) has been created and the planning process can stop. Second, in every cycle, the algorithm uses the internal state to determine which goals need to be planned for and which have already been achieved by following a means-ends analysis strategy. Unlike some other planners which analyze all of the possible effects of the operators that may have changed the initial state, FLECS simply checks if a particular goal is true in the current state.[5] Third, the planner uses the state to determine which operators may now be applied: i.e., those whose preconditions are all true in the state. Fourth, FLECS can use its state to choose an operator and bindings that are most likely to achieve a particular goal with a minimum of planning effort (Blythe & Veloso, 1992). In summary, and with reference to the algorithm in Table 1, FLECS uses the state to determine:

- if the goal statement has been satisfied (step 2);
- which goals still need to be achieved (step 3);
- which operators are applicable (step 3);
- which operators to try first while planning (step 6).

In planners that do not keep an internal state, all four of these steps require considerable planning effort when they are even attempted at all. In contrast, FLECS can perform these steps in sub-quadratic time. Furthermore, other planners do not have any particular methods for choosing among possible operators to achieve a goal. This particular use of state has been shown to provide significant efficiency gains in PRODIGY4.0 (Veloso & Blythe, 1994).

Since FLECS *does* use the state, it makes a big difference whether or not it chooses to change its state (*apply* an operator) at a given time. The advantage of applying an operator is that more informed planning results during each of the above four steps. However, the choice to apply an operator involves a commitment to order this operator before all other operators that have not yet been applied. This commitment is only temporary since if no plan can be found with the operator in this position, the operator can be "un-applied" by simply changing the internal state back to its previous status. One may argue that the requirement that operators be applied in an explicit order opens up the possibility of exponential backtracking. However this argument is vacuous, as planning is undecidable (Chapman, 1987). Due to the use of state, FLECS can reduce the likelihood of requiring backtracking at the operator choice point. In so doing, it may increase the likelihood of backtracking at the operator-ordering choice point. However, it has the flexibility of being able to come down on either side of this tradeoff.

5. Note that since the goal and the state are fully instantiated, this matching can be accomplished in constant time for each goal by using a hash table of literals.

### 2.2 The Choice of Commitment Strategies

In order to control the tradeoff between eager and delayed state changes, FLECS has a toggle which determines whether the algorithm prefers subgoaling or applying an operator in step 5. Which option FLECS considers first may affect its path through the search space and consequently its planning efficiency. This ability to accommodate different types of search is the most novel part of our algorithm. Its significance lies in the difference between *subgoaling* and *applying*.

The difference between subgoaling and applying is illustrated in Figure 1. Subgoaling can be best understood as regressing one goal, or backward chaining, using means-ends analysis. It includes the choices of a goal to plan for and an operator to achieve this goal. As seen in Section 2.1, both of these choices are affected by FLECS's internal state. Thus, subgoaling without ever updating the internal state (applying an operator) can lead to uninformed planning decisions. On the other hand, by subgoaling extensively, FLECS can select a large set of operators that will appear in the plan before deciding in which order to apply them. Then FLECS takes into account the conflicts, or "threats," among operators and orders them appropriately when applying them.

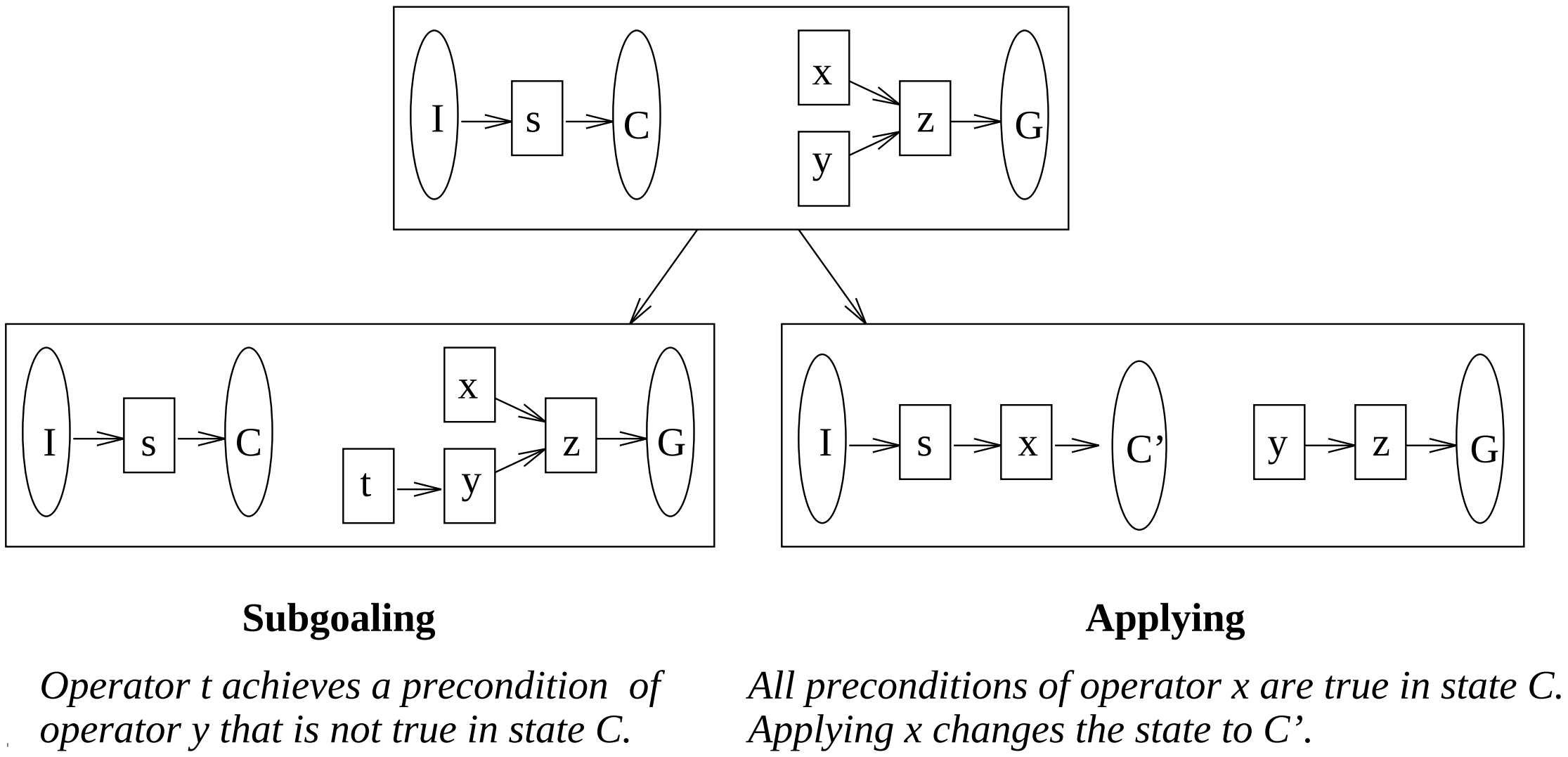

Figure 1: This diagram from (Fink & Veloso, 1994) illustrates the difference between subgoaling and applying. A search node consisting of a "head-plan" and a "tail-plan." The head-plan contains operators that have already been applied and have changed the initial state "I" to the current state "C." The tail-plan consists of operators that have been selected to achieve goals in the goal statement "G" and operators that have been selected to achieve preconditions of these operators, etc. The figure shows how the planner could either subgoal or apply at a given search node.

Applying an operator is FLECS's way of changing the current internal state so that future subgoaling decisions can be more informed. However, applying an operator is a commitment (temporary since backtracking is possible) that this operator should be executed before any other. This is the essential tradeoff between eagerly subgoaling and eagerly applying: eagerly subgoaling delays ordering commitments (delayed commitment), while eagerly applying facilitates more informed subgoaling (eager commitment).

flecs has a switch (toggle) that can change its behavior from eager subgoaling to eager applying and vice versa at any time. This feature is the most significant improvement in flecs over prodigy4.0 and its predecessors. Since we saw evidence that neither delayed-commitment nor eager-commitment search strategies were consistently effective (Stone et al., 1994), we felt the need to provide flecs with the toggle. Thus, flecs can combine the advantages of delayed commitments and eager commitments.[6]

## 3. An Illustrative Example

In this section we present an example that illustrates in detail most of the planning situations that can arise in a general planning problem. Although planning may be well understood in general, past descriptions of planning algorithms have not directly addressed most of these situations in full detail. The flecs algorithm is designed to handle all of these situations.

In order to describe flecs completely, we need to define several variables that are maintained as the algorithm proceeds. Since it is much easier to understand the algorithm once one is familiar with the concepts that these variables denote, we present an annotated example in Figures 2 through 9 before formally presenting flecs. We further recommend following how each of the variables and functions $\mathcal{C}$, $\mathcal{G}$, $\mathcal{P}$, $\mathcal{O}$, $\mathcal{A}$, $a$, and $c$ change throughout the annotated example, according to their definitions:

- $\mathcal{C}$ represents the *current internal state* of the planner. Its uses are summarized in Section 2.1.
- $\mathcal{G}$ is the set of *goals and subgoals* that the planner is aiming to achieve. These are the goals that are on the *fringe* of the subgoal tree. Goals in $\mathcal{G}$ may be goals that have not yet been planned for, or goals that have been achieved (perhaps trivially) but not yet used by the operator that needs them as one of its preconditions (i.e., this operator has not been applied yet).
- $\mathcal{P}$ is the set of *pending goals*: goals in $\mathcal{G}$ that may need to be planned for in the current state.
- $\mathcal{O}$ stands for the set of *instantiated operators* that have been selected to achieve goals and subgoals.
- $\mathcal{A}$ is the set of *applicable operators*: operators in $\mathcal{O}$ whose preconditions are all satisfied in the current state and which are needed in the current state to achieve some goal.
- For a goal $G$, $a(G)$ is the set of its *ancestor goal sets* – the sequences of goals that caused $G$ to become a member of $\mathcal{G}$. Trivially, a goal is an ancestor of each of the preconditions of the operator selected to achieve this goal. $a(G)$ is a set of sets because G can have different sets of ancestors. This concept will become clearer through the example.

6. In Section 5 we discuss different heuristics to guide this choice and we discuss our view of toggle as a perfect focus for learning.

- For an operator $O$, $c(O)$ is the set of goals which $O$ was selected to achieve – its *causes*. Applying $O$ establishes each member of $c(O)$. As illustrated below, the functions $a$ and $c$ are needed to determine which goals are pending and which operators are applicable. They are analogous to causal links used to determine threats in other planners (Chapman, 1987; McAllester & Rosenblitt, 1991).

The sequence of planning decisions in this example (Figure 2 through Figure 9) is designed to illustrate the uses of all of FLECS's variables and functions. We recommend becoming familiar with them by spending some time carefully tracing their values and returning to the above definitions throughout this example. Note that the figures show only the tail-plan while we mention applied operators and state changes in the text. Goals are in circles: solid circles if they are not true and dashed circles if they are true in the current state. Operators are in boxes with arrows pointing to the goals which they "produce," i.e., the goals which the operators have been selected to achieve (their causes). In turn, the preconditions of these operators are goals with arrows pointing to the operators which "consume" them. Operators that are applicable in the current state appear in bold boxes. Changes to the functions $c$ and $a$ are underlined in the captions.

We present now the example. Figure 2 shows the initial planning situation, in which we consider a planning problem with three literals in the goal statement, $G_1$, $G_2$, and $G_3$, i.e., $\mathcal{G} = \{G_1, G_2, G_3\}$. There is one literal in the initial state, $G_7$, i.e., $\mathcal{C} = \{G_7\}$. As none of the goals is true in the initial state, $\mathcal{P} = \mathcal{G}$. There are no operators selected, i.e., $\mathcal{O} = \emptyset$, and therefore also no operators applicable, i.e., $\mathcal{A} = \emptyset$. At this point, since they are all top-level goals, none of the goals has any ancestors: $\underline{a(G_1) = a(G_2) = a(G_3) = \emptyset}$. As there are no applicable operators, the next step must be to subgoal on one of the pending goals.

$\mathcal{C} = \{G_7\}$
$\mathcal{G} = \{G_1, G_2, G_3\}$
$\mathcal{O} = \emptyset$
$\mathcal{P} = \{G_1, G_2, G_3\}$
$\mathcal{A} = \emptyset$

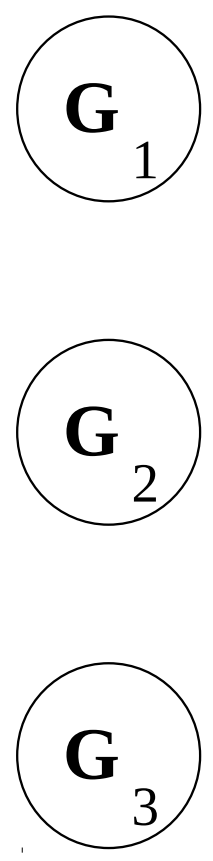

Figure 2: An example: The initial specification of a planning situation.

Figure 3 shows the planning situation after FLECS subgoals on $\boldsymbol{G_1}$ and $\boldsymbol{G_2}$. Suppose that operator $O_1$, with preconditions $G_6$ and $G_7$, is selected to achieve $G_1$, while $O_2$ is chosen to achieve $G_2$ as indicated below. Note that the operators' preconditions replace their causes in the set of fringe goals $\mathcal{G}$; since $G_7$ is true in the current state, it is NOT included in the set of pending goals $\mathcal{P}$. Here $G_1$ is the *cause* of $O_1$, so $\underline{c(O_1) = \{G_1\}}$; similarly, $c(O_2) = \{G_2\}$. The new goals all have nonempty ancestor sets: $a(G_6) = a(G_7) = \{\{G_1\}\}$, and $a(G_4) = \{\{G_2\}\}$. There are still no applicable operators: $O_1$ cannot be applied because $G_6 \notin \mathcal{C}$ and $O_2$ cannot be applied because $G_4 \notin \mathcal{C}$. Therefore, FLECS subgoals again.

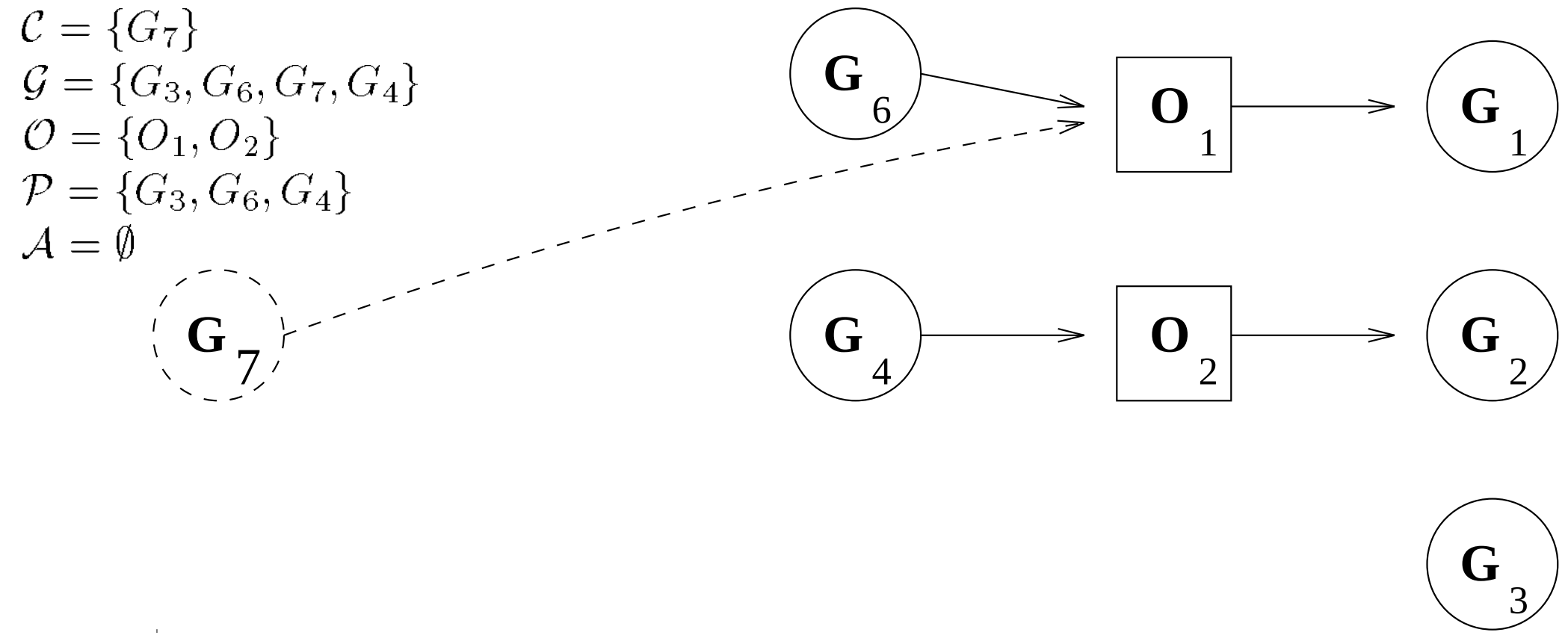

Figure 3: Resulting planning situation after subgoaling on $\boldsymbol{G}_1$ and $\boldsymbol{G}_2$.

Figure 4 shows the planning situation after FLECS subgoals on $\boldsymbol{G}_3$. Suppose that the operator selected to achieve $G_3$ has preconditions $G_4$ and $G_5$. We now have $c(O_3) = \{G_3\}$, and $a(G_5) = \{\{G_3\}\}$. The causes of operators $O_1$ and $O_2$ do not change, so $c(O_1) = \{G_1\}$ and $c(O_2) = \{G_2\}$ as in the previous step. Similarly, $a(G_6)$ and $a(G_7)$ remain unchanged. However, $G_4$ now has *two* sets of ancestor goals: $a(G_4) = \{\{G_2\}, \{G_3\}\}$. To understand the need to keep both ancestor sets, consider the possibility that $G_2$ could be achieved unexpectedly as a side-effect of some unrelated operator instead of being achieved by $O_2$ as planned for. In this case, $G_4$ would remain a pending goal since it would be needed to achieve $G_3$. Again, since there are no applicable operators, FLECS must subgoal on one of the pending goals, i.e., $G_6$, $G_4$, or $G_5$.

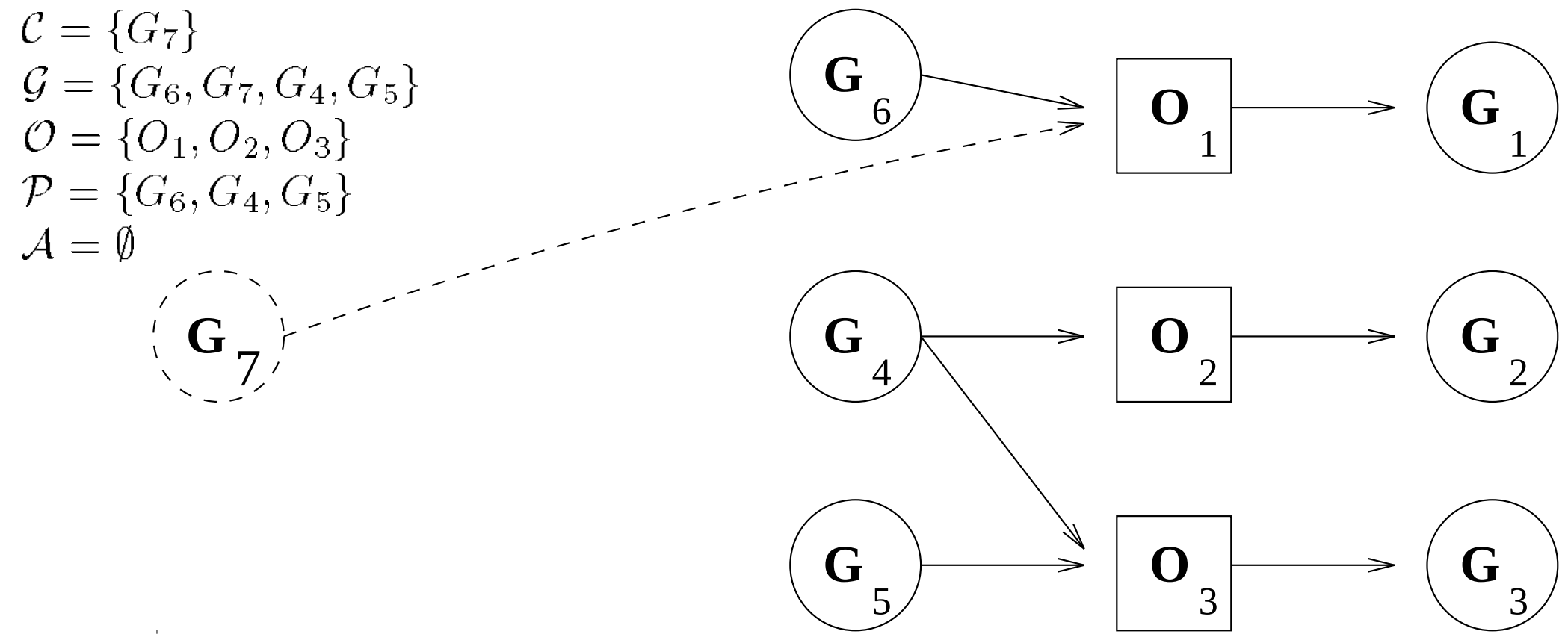

Figure 4: Resulting planning situation after subgoaling on $\boldsymbol{G}_3$.

Figure 5 shows the planning situation after FLECS subgoals on $\boldsymbol{G}_4$. Suppose that $O_4$ — an operator with precondition $G_7$ — is selected to achieve $G_4$. Since $G_7$ is true in the current state, $O_4$ is our first applicable operator. Note that it is necessarily ordered before $O_2$ and $O_3$ since its cause is a precondition of these operators. As usual, the cause of the new operator is stored: $c(O_4) = \{G_4\}$. In addition, the ancestors of $G_7$ must be augmented to include two new ancestor sets: $a(G_7) = \{\{G_1\}, \{G_4, G_2\}, \{G_4, G_3\}\}$. Although there is now an applicable operator, let us assume that FLECS chooses to delay its commitment to order $O_4$ as the first step in the plan and subgoals again on a pending goal.

$\mathcal{C} = \{G_7\}$
$\mathcal{G} = \{G_6, G_7, G_5\}$
$\mathcal{O} = \{O_1, O_2, O_3, O_4\}$
$\mathcal{P} = \{G_6, G_5\}$
$\mathcal{A} = \{O_4\}$

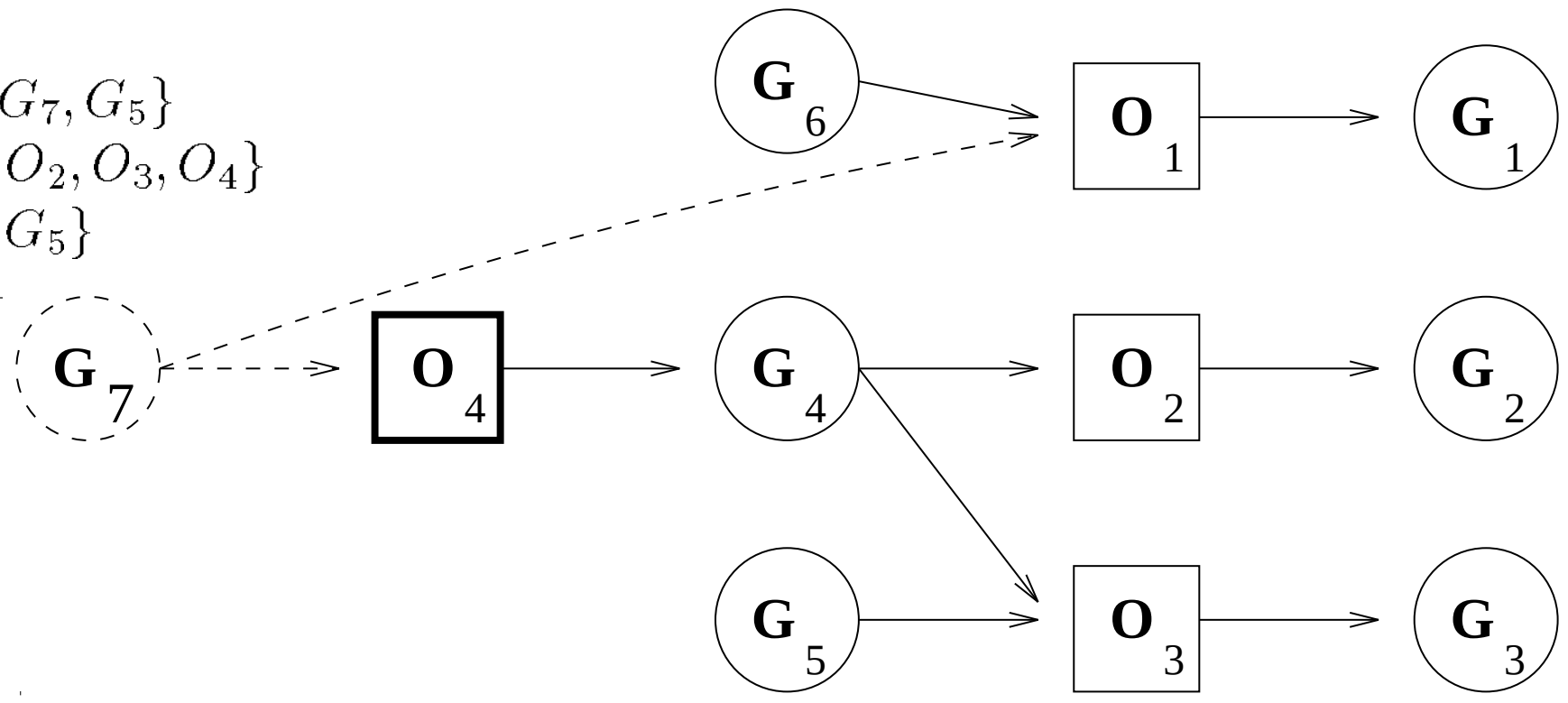

Figure 5: Resulting planning situation after subgoaling on $\boldsymbol{G}_4$.

Figure 6 shows the planning situation after FLECS subgoals on $\boldsymbol{G}_5$. Suppose that operator $O_4$ can also achieve $G_5$ and that it is selected to do so. We now need to update both the causes of this operator and the ancestors of its precondition: $c(O_4) = \{G_4, G_5\}$ and $a(G_7) = \{\{G_1\}, \{G_4, G_2\}, \{G_4, G_3\}, \{G_5, G_3\}\}$. Now rather than subgoaling on the last remaining pending goal ($G_6$), let us apply $O_4$. Note that this decision corresponds to an early commitment in terms of ordering the operators $O_1$, $O_4$, and any operators later selected to achieve $G_6$ which are unordered by the current planning constraints. FLECS changes here from its delayed-commitment strategy to an eager-commitment strategy.

$\mathcal{C} = \{G_7\}$
$\mathcal{G} = \{G_6, G_7\}$
$\mathcal{O} = \{O_1, O_2, O_3, O_4\}$
$\mathcal{P} = \{G_6\}$
$\mathcal{A} = \{O_4\}$

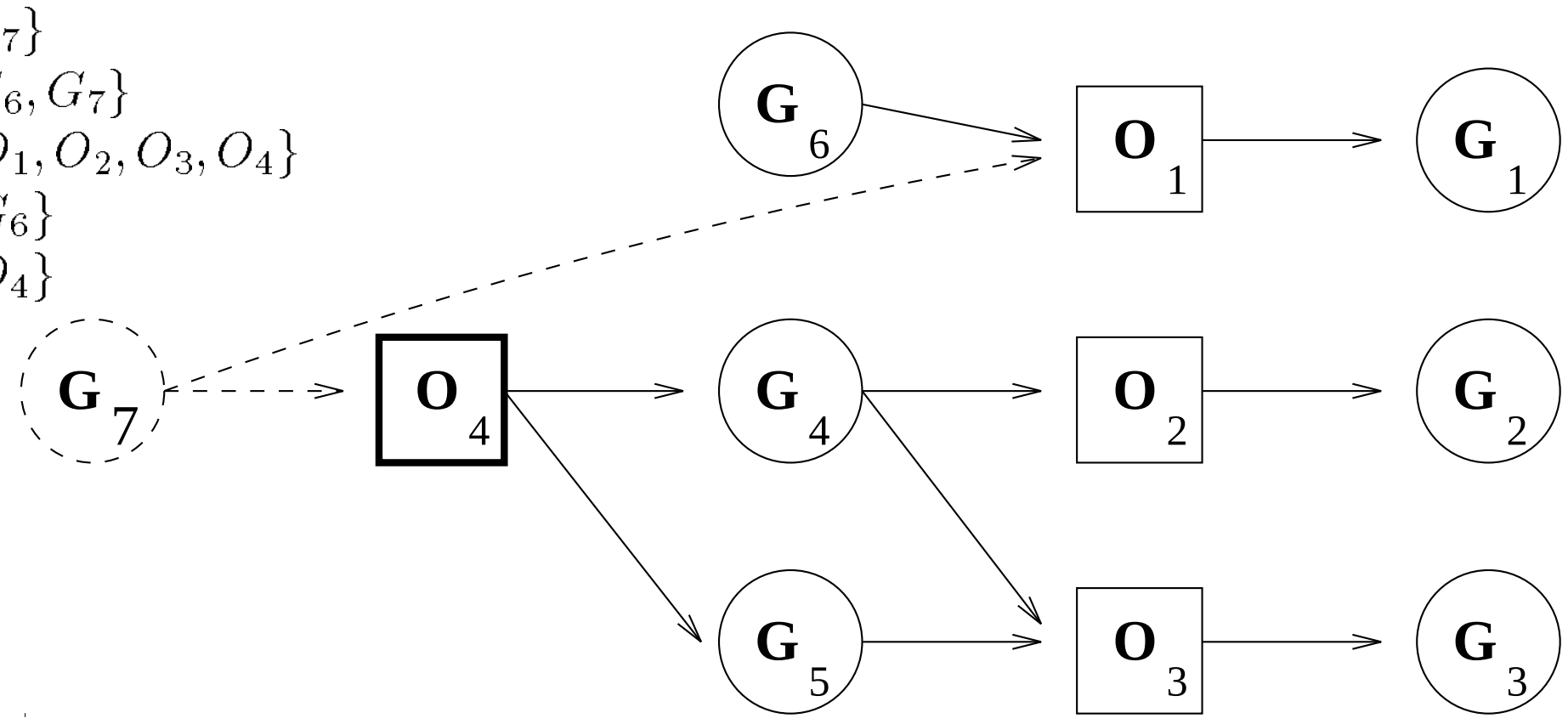

Figure 6: Resulting planning situation after subgoaling on $\boldsymbol{G}_5$.

Figure 7 shows the planning situation after FLECS applied $\boldsymbol{O}_4$. Since operator $O_4$ was applied in order to achieve goals $G_4$ and $G_5$, they are both true in the current state and back on the fringe of the goal tree, i.e., they are in $\mathcal{C}$ and $\mathcal{G}$. Notice that they stay in $\mathcal{G}$ until eventually they have been "consumed" by $O_2$ and $O_3$. However, since they are true in the current state, they are not pending goals. Since $G_7$ is once again the precondition of only one selected operator, $\underline{a(G_7) = \{\{G_1\}\}}$ as before. $O_2$ and $O_3$ are now applicable as their preconditions are all true in the current state thanks to $O_4$. Let us assume that FLECS maintains the eager-commitment strategy and continues applying applicable operators. FLECS orders $O_2$ before $O_3$, since $O_3$ deletes a precondition of $O_2$ (effects are not shown).

$\mathcal{C} = \{G_7, G_4, G_5\}$
$\mathcal{G} = \{G_6, G_7, G_4, G_5\}$
$\mathcal{O} = \{O_1, O_2, O_3\}$
$\mathcal{P} = \{G_6\}$
$\mathcal{A} = \{O_2, O_3\}$

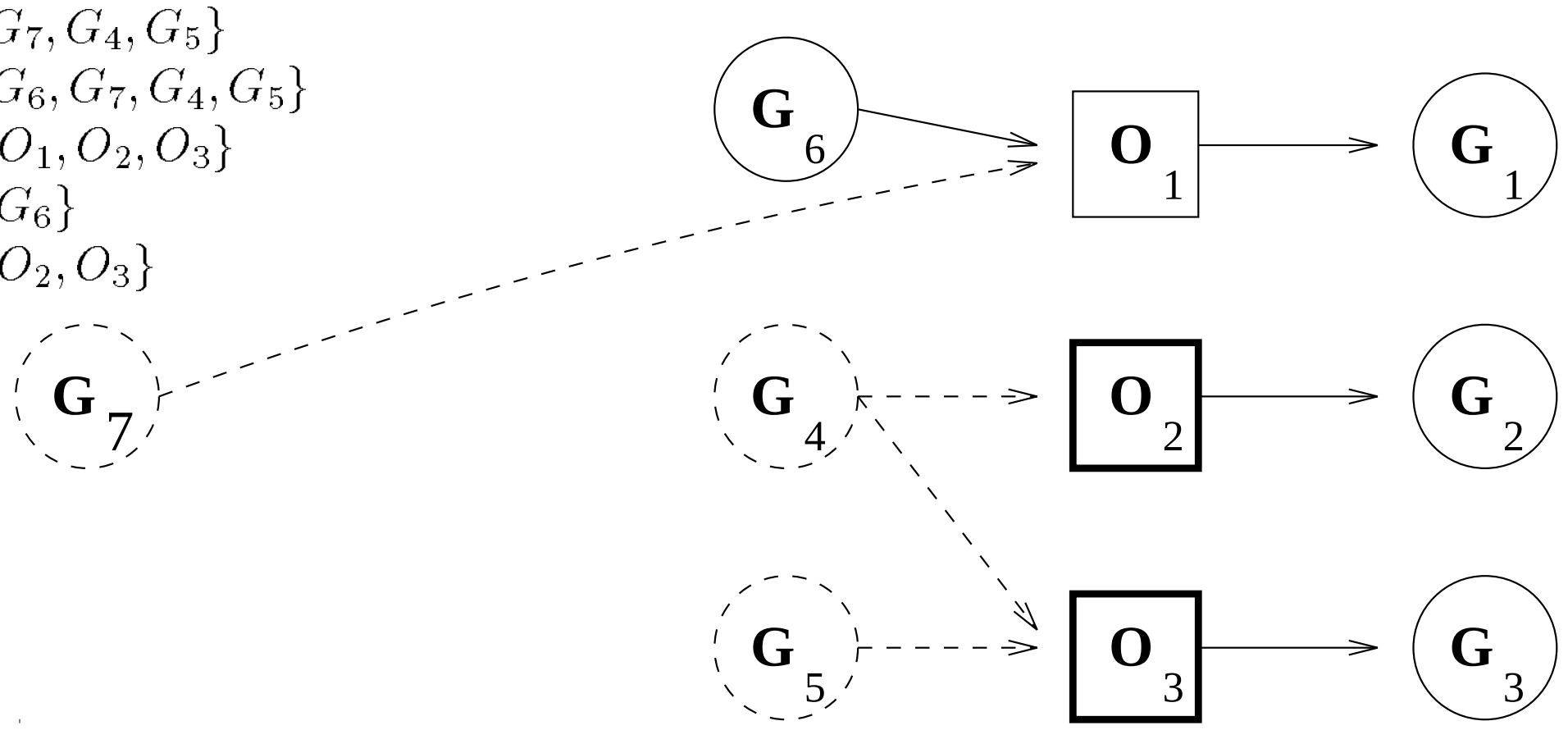

Figure 7: Resulting planning situation after applying $\boldsymbol{O}_4$ from Figure 6.

Figure 8 shows the planning situation after FLECS applied $\boldsymbol{O}_2$. Suppose that, although it was not selected to do so, operator $O_2$ achieves $G_1$ as a side-effect. Perhaps $O_2$ has a conditional effect that was not visible to the planner, or perhaps $O_1$ simply looked more promising than $O_2$ as an operator to achieve $G_1$ at the time when it was selected. In any

$\mathcal{C} = \{G_7, G_4, G_5, G_1, G_2\}$
$\mathcal{G} = \{G_6, G_7, G_4, G_5, G_2\}$
$\mathcal{O} = \{O_1, O_3\}$
$\mathcal{P} = \emptyset$
$\mathcal{A} = \{O_3\}$

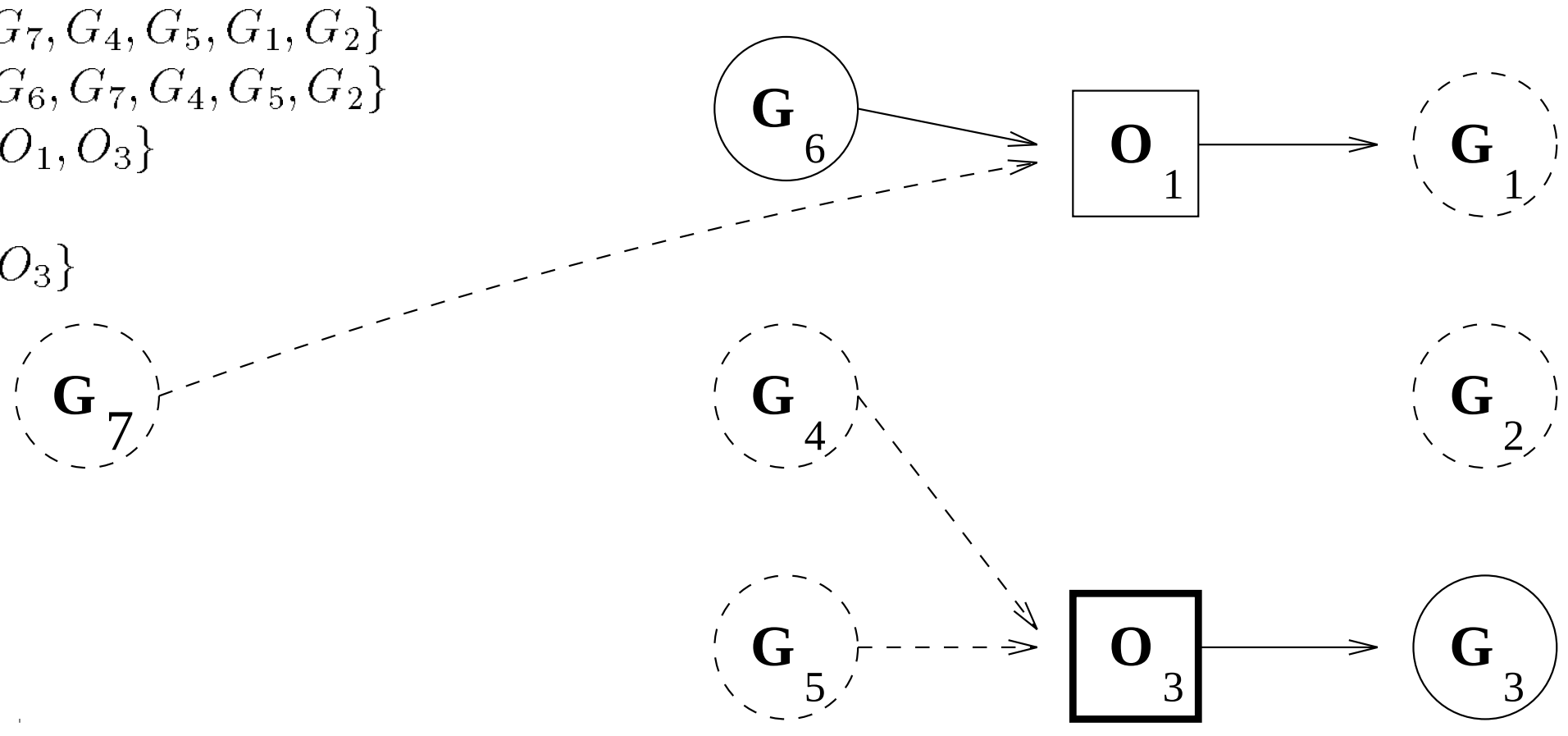

Figure 8: Resulting planning situation after applying $\boldsymbol{O}_2$ from Figure 7.

case, $G_1$ is now in $\mathcal{C}$ and the planning done for it is no longer needed: $G_6$ is no longer a pending goal, since its sole ancestor is already in $\mathcal{C}$. This fortuitous achievement of a goal is the reason that we need to use the functions $c$ and $a$ to adjust the sets of pending goals $\mathcal{P}$ and applicable operators $\mathcal{A}$: it would be wasted effort for FLECS to plan to achieve $G_6$. Note that were $G_6$ a precondition of $O_3$ as well as $O_1$, it *would* be a pending goal since it would still be relevant to achieving $G_3$. At this point, only the ancestors of $G_4$ must be reset: $\underline{a(G_4) = \{\{G_3\}\}}$. Since there are no more pending goals, FLECS must now apply the last remaining applicable operator, $O_3$.

Figure 9 shows the final planning situation after FLECS applied $\boldsymbol{O_3}$. At this point all of the top level goals are true in the current state. Despite the fact that some of the planning tree remains, FLECS recognizes that there is no more work to be done and terminates. The final plan is $O_4$, $O_2$, $O_3$, which is the sequence of operators applied in the head-plan (not shown) corresponding to the steps in Figures 7, 8, and 9. An a posteriori algorithm (Veloso, Pérez, & Carbonell, 1990) can convert the sequence into a partially ordered plan capturing the dependencies: $O_4$, $\{O_2, O_3\}$.

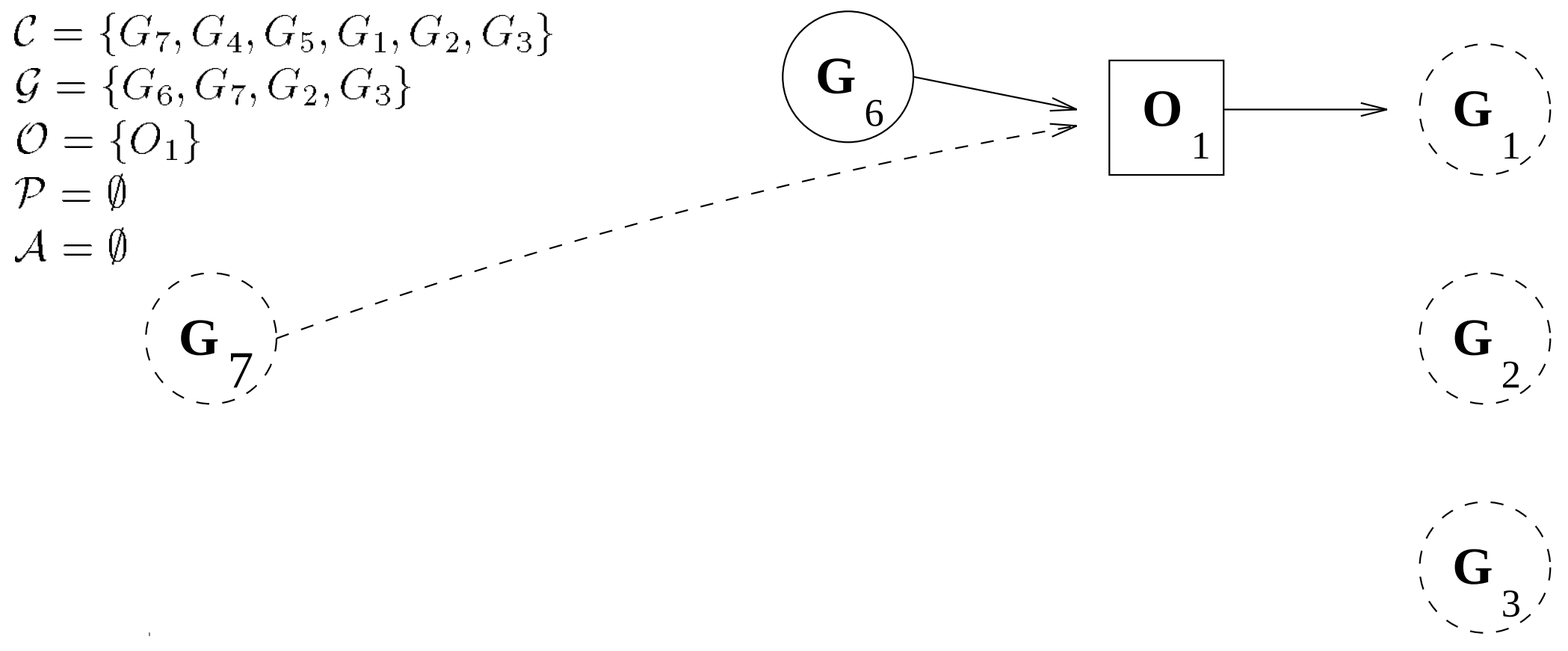

Figure 9: Final planning situation after applying $\boldsymbol{O_3}$ from Figure 8.

## 4. FLECS: The Detailed Description

Aside from the variables and functions introduced in the preceding section, we need to define only four more things before presenting the complete algorithm. First, *Initial State* and *Goal Statement* are the corresponding ground literals from the problem definition. Second, for a given operator $O$, $pre(O)$, $add(O)$, and $del(O)$ are its instantiated preconditions, add-list, and delete-list respectively. FLECS takes these values straight from the domain representation, which may include disjunctions, negations, existentially and universally quantified preconditions and effects, and conditional effects (Carbonell et al., 1992). When $O$ has conditional effects, $add(O)$ and $del(O)$ are determined dynamically, using the state at the time $O$ is applied. Third, the "*relevant instantiated operators* that could achieve $G$" (step 6) are all the instantiated operators $O$ (operators with fully-specified bindings) which have $G \in add(O)$ if $G$ is a positive goal or $G \in del(O)$ if $G$ is a negative goal. Fourth, *toggle* is a variable that determines the flavor of search, as described later.

### 4.1 The Planning Algorithm

We present the FLECS planning algorithm in full detail in Table 2.[7] While examining the algorithm, notice that the fringe goals $\mathcal{G}$, the selected operators $\mathcal{O}$, the ancestor function $a(G)$, the cause function $c(O)$, and the current state $\mathcal{C}$ are maintained incrementally. On the other hand, the pending goals $\mathcal{P}$, the applicable operators $\mathcal{A}$, and *toggle* are recomputed on every pass through the algorithm.

Step 1 initializes most of these variables. At the beginning of the planning process, the only goals in $\mathcal{G}$ are those in the goal statement, the current state $\mathcal{C}$ is the same as the initial state, and since no operators have yet been selected, $\mathcal{O}$ is empty. Both the ancestor function $a$ and the cause function $c$ are initialized to the constant function that maps everything to $\emptyset$. In practice, the domain of $a$ is the set of goals and the domain of $c$ is the set of operators that appear in the problem. However, since most of these goals and all of these operators have not been determined when the algorithm is first called, we must initialize the functions with unrestricted domains.

Step 2 is the termination condition. It is called after each time a new operator is applied. The algorithm terminates successfully if every goal $G$ in the goal statement is true, or satisfied, in the current state $\mathcal{C}$, i.e., $G \in \mathcal{C}$.

In step 3, the sets of pending goals and applicable operators are computed based on the current state. Pending goals are the goals that the planner may need to plan for. Initially, the pending goals are the fringe goals that are not currently true or that were true in the initial state.[8] The applicable operators are the selected operators whose preconditions are true in the state.

Then, step 4 computes the pending goals $\mathcal{P}$ and applicable operators $\mathcal{A}$ that are *active* in the current state. A pending goal is active as long as it is on the fringe of the subgoal tree and it still needs to be planned for. A goal is no longer active if every one of its ancestor sets has at least one goal that has already been achieved: then all purposes for which the goal was selected no longer exist (as was the case for $G_6$ in Figure 8). An applicable operator is active in the current state as long as it would achieve a goal that is still useful to the plan. An applicable operator is no longer active if each of its causes is either true in the current state or no longer active.

Step 5 is the most novel part of our algorithm. It allows for a flexible search strategy within a single planning algorithm. Since at this step, FLECS has not yet terminated, there must be either some active pending goals or active applicable operators, i.e., $\mathcal{P}$ or $\mathcal{A}$ must be non-empty. However, if there is only one or the other, then there is no choice to be made. If, on the other hand, both $\mathcal{P}$ and $\mathcal{A}$ are non-empty, then we can either proceed to step 6 or to step 7. For the sake of completeness, we must keep both options open; but which option FLECS considers first may affect the amount of search required. By changing

7. The detail of this algorithm allows the reader to carefully study and re-implement FLECS.
8. Since the planner cannot backtrack beyond the initial state, we must keep goals from the initial state as pending goals for the sake of completeness.

| | |
|---|---|
| $\mathcal{C}$: | *current state* |
| $\mathcal{G}$: | *fringe goals* |
| $\mathcal{P}$: | *pending goals* |
| $\mathcal{O}$: | *instantiated operators* |
| $\mathcal{A}$: | *applicable operators* |
| $a$: | *ancestor goal sets* |
| $c$: | *causes* |

1. Initialize:
   a. $\mathcal{G}$ = Goal Statement.
   b. $\mathcal{C}$ = Initial State.
   c. $\mathcal{O} = \emptyset$.
   d. $\forall G.a(G) = \emptyset$.
   e. $\forall O.c(O) = \emptyset$.
2. Terminate if Goal Statement $\subseteq \mathcal{C}$.
3. Compute applicable operators $\mathcal{A}$ and pending goals $\mathcal{P}$:
   a. $\mathcal{P} = \{G \in \mathcal{G} \mid G \notin \mathcal{C} \vee G \in$ Initial State$\}$.
   b. $\mathcal{A} = \{A \in \mathcal{O} \mid pre(A) \subseteq \mathcal{C}\}$.
4. Adjust $\mathcal{P}$ and $\mathcal{A}$ to contain only active members:
   a. $\mathcal{P} = \mathcal{P} - \{P \in \mathcal{P} \mid \forall S \in a(P).\exists G \in S \text{ s.t. } G \in \mathcal{C}\}$.
   b. $\mathcal{A} = \mathcal{A} - \{A \in \mathcal{A} \mid \forall G \in c(A).[(G \in \mathcal{C}) \vee (\forall S \in a(G).\exists G' \in S \text{ s.t. } G' \in \mathcal{C})]\}$.
5. Subgoal or Apply:
   a. **Set or reset toggle to** *sub* **or** *app*, i.e. Set default to delayed or eager commitment.
   b. If $\mathcal{A} = \emptyset$, go to step 6.
   c. If $\mathcal{P} = \emptyset$, go to step 7.
   d. Choose to apply or to subgoal (*backtrack point*):
      - If **toggle** = $sub \wedge P \not\subseteq \mathcal{C}$, subgoal first: go to step 6.
      - If **toggle** = *app*, apply first: go to step 7.
6. Choose a goal $P$ from $\mathcal{P}$ (*not a backtrack point*).
      - Choose a goal not true in the Current State using means-ends analysis.
   a. Get the set $\mathcal{R}$ of relevant instantiated operators that could achieve $P$.
   b. If $\mathcal{R} = \emptyset$ then
      i. $\mathcal{P} = \mathcal{P} - \{P\}$.
      ii. If $\mathcal{P} = \emptyset$ then fail (i.e., backtrack).
      iii. Go to step 6.
   c. Choose an operator $O$ from $\mathcal{R}$ (*backtrack point*).
      - Choose the operator with minimum conspiracy number, i.e. the operator which appears to be achievable with the least amount of planning.
   d. $\mathcal{O} = \mathcal{O} \cup \{O\}$.
   e. $\mathcal{G} = (\mathcal{G} - \{P\}) \cup pre(O)$.
   f. $c(O) = c(O) \cup \{P\}$.
   g. $\forall G \in pre(O).a(G) = a(G) \cup \{\{P\} \cup S \mid S \in a(P)\}$.
   h. Go to step 3.
7. Choose an operator $A$ from $\mathcal{A}$ (*backtrack point for interactions*).
      - Use a heuristic to find operators with fewer interactions – similar to the one used by the SABA heuristic.
   a. Apply $A$: $\mathcal{C} = (\mathcal{C} \cup add(A)) - del(A)$
   b. $\mathcal{O} = \mathcal{O} - \{A\}$.
   c. $\forall G \in pre(A).a(G) = a(G) - \{S \in a(G) \mid S \cap c(A) \neq \emptyset\}$.
   d. $\mathcal{G} = (\mathcal{G} \cup c(A)) - \{G \in pre(A) \mid a(G) = \emptyset\}$.
   e. $c(A) = \emptyset$.
   f. Go to step 2.

Table 2: The full description of FLECS.

the value of *toggle*, which can be done on any pass through the loop, FLECS can change the type of search as it works on a problem.

Each pass through the body of the algorithm visits either step 6 or step 7. When subgoaling (step 6), an active pending goal $P$ is chosen from $\mathcal{P}$. Note that unlike the corresponding choice in step 7, this choice of subgoals is not a backtrack point. However, if there are no operators that could achieve this goal, then another goal is chosen (step 6b). Means-ends analysis is used as a heuristic to prefer subgoaling on goals that are not currently true. Next, an operator $O$ is chosen that could achieve the chosen goal (step 6c). It can either be a new operator or an existing one as in Figure 6 ($O_4$, which had already been selected to achieve $G_4$, is also selected to achieve $G_5$). The choice of operator is a backtrack point. Unless some other heuristic is provided, the minimum conspiracy number heuristic is used to determine which operator should be tried first (Blythe & Veloso, 1992). In short, this heuristic selects the instantiated operator that appears to be achievable with the least amount of planning.

Before returning to the top of the loop, all of the affected variables are updated. First, $O$ is added to $\mathcal{O}$ using set union so that the same operator never appears twice (step 6d). Second, $O$'s preconditions are added to $\mathcal{G}$, while $P$ is removed (step 6e): once $P$ has an operator selected to achieve it, it is no longer on the fringe of the subgoal tree. Third, the cause of $O$ is augmented to include $P$ (step 6f). Fourth, the ancestor sets of $O$'s preconditions are augmented to include all sets of goals comprised of $P$ and *its* ancestors (step 6g). As explained in Figure 4, all ancestor sets must be included. Finally, since the state is not changed at all, the termination condition cannot be met. The algorithm returns to step 3.

When applying an operator (step 7), an applicable operator $A$ is chosen from $\mathcal{A}$. A heuristic that analyzes the applicable operators can be used to choose the best possible operator. One such heuristic analyzes interactions between operators by identifying negative threats, similarly to the SABA heuristic in (Stone et al., 1994). In short, this heuristic prefers operators that do not delete any preconditions of, and whose effects are not deleted by, other operators. This choice of an applicable operator is a backtrack point where all orderings of interacting applicable operators are considered. Different orderings of completely independent operators need not be considered. Completely independent operators are those with interactions neither between themselves nor among their ancestor sets. Since the application of one such operator can make no difference to the application of another, we only need to consider one ordering of these operators.

Once $A$ is chosen, it is promptly applied (step 7a). This application involves changing the current state as prescribed by $A$. Note that if $A$ has conditional effects, they are expanded at this point. Next, the relevant variables are updated. First, updating involves removing $A$ from the set of selected operators (step 7b). Second, the ancestors of $A$'s preconditions are only those ancestor sets which did not include $A$ (step 7c): $A$ does not need further planning. Figure 7 shows an example in which a precondition ($G_7$) does still have an ancestor remaining. Third, since $A$ has been applied, its preconditions that are not goals for any other reason are no longer on the fringe, but its causes are (step 7d): if they are unachieved they must be re-achieved. Fourth, in case $A$ is ever selected again as an operator to achieve some goal, $c(A)$ is reset to $\emptyset$ (step 7e). Finally, since the current state has been altered, the algorithm returns to step 2 where the termination condition is checked.

### 4.2 Discussion: Backtracking, Heuristics, and Properties

One should pay close attention to the placement of backtrack points in the algorithm. In particular, there are only three: the subgoal/apply choice in step 5, the choice of operator to achieve a goal in step 6, and the choice of applicable operator in step 7. However, the choice of goal on which to subgoal in step 6, which is a backtrack point in the PRODIGY algorithm, is not a backtrack point here. FLECS does not need this backtrack point because the choice to apply or not to apply an operator at a given time is left open in step 5 and all significantly different orders of applying applicable operators are considered in step 7. As explained in the previous subsection, different orderings of completely independent operators are not considered. Nevertheless, all orderings that could lead to a solution are considered. Therefore, backtracking on the choice of subgoal would only cause redundant search. This elimination of a backtrack point is a significant improvement in FLECS over previous implementations, namely NoLimit and PRODIGY4.0. Note that no new backtrack points are added to offset the eliminated backtrack point.

FLECS's only explicit failure point is in step 6 and occurs when the algorithm has chosen to subgoal, but none of the pending goals have any relevant operators. All other failures are implicit. That is, at a backtrack point, if all choices have been unsuccessfully tried then the algorithm backtracks. As presented, the algorithm only terminates unsuccessfully if the entire search space has been exhausted. Other causes for failure, such as goal loops, state loops, depth bounds, and time limits, are incorporated in the same manner as in PRODIGY4.0 (Carbonell et al., 1992).

At each choice point, there is some heuristic to determine which branch to try (first). In step 6, the goal is chosen using means-ends analysis, and the operator with the minimum conspiracy number is chosen to achieve that goal. In step 7, the choice mechanism from the SABA heuristic is used to determine which applicable operator to try first. In step 5, *toggle*, which can be changed at any time, determines whether the default commitment strategy should be eager subgoaling or eager applying. Note that if all of the pending goals are true in the Current State (or if there are no pending goals), the planner may apply an applicable operator regardless of the value of *toggle*. Similarly, if there are no applicable operators, the planner must subgoal even if *toggle* indicates to prefer applying. *toggle* is a new variable to guide heuristic search at an existing choice point with a branching factor of two: it does not represent the addition of a new backtrack point. As discussed throughout, it provides FLECS with the ability to change its commitment strategy. As suggested by its name, *toggle* can be one of two values: *sub* and *app* indicating eager subgoaling and eager applying respectively.

Here we describe a domain-independent heuristic that could be used to guide changes to the value of *toggle*. Such a heuristic should allow eager commitments when there is reason to believe that there will not be a need to backtrack over the resulting operator linearization. In this case, setting *toggle* to *app* will increase the planning efficiency by converting a partially-ordered set of operators into a sequence that leads to a single possible state, which can then be used to guide subsequent planning. This process is equivalent to starting a new and smaller planning problem as all the previous choices will be embedded in the state.

The situation described above is similar to that which arises in the ALPINE system which constructs efficient abstraction hierarchies (Knoblock, 1994). ALPINE can guarantee that planning hierarchically using its generated abstraction hierarchies will not lead to backtracking across refinement spaces. Figure 10 illustrates how FLECS can use this abstraction planning information to control the value of *toggle*. If *toggle* changes to *app* when a particular abstract planning step is completely refined and the abstraction hierarchies preserve ALPINE's ordered monotonicity property, then there should be no need to backtrack over the resulting operator ordering. Then *toggle* can change back to *sub*, and FLECS can continue planning with updated state information.

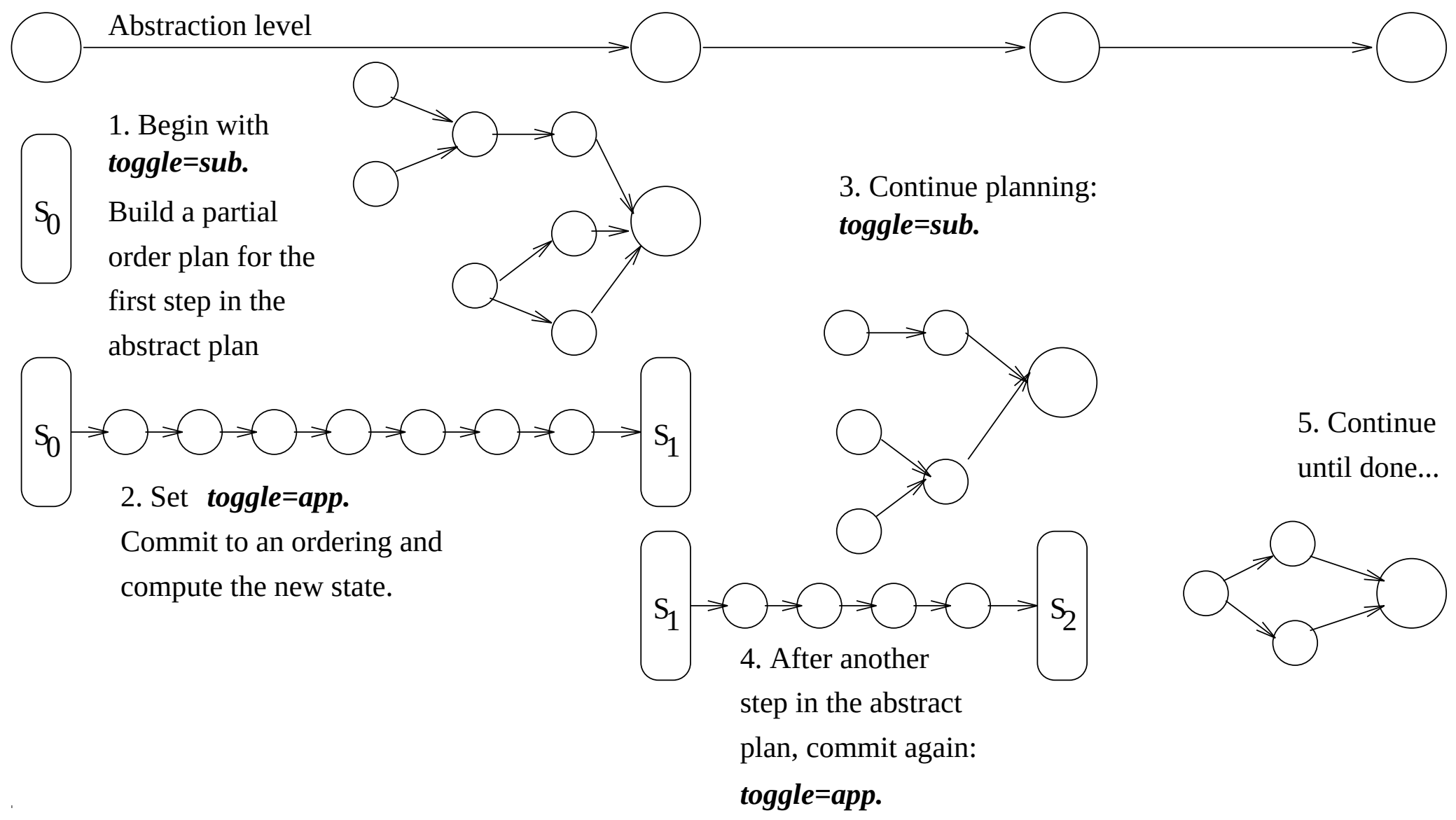

Figure 10: Using abstraction information to guide changes to *toggle*.

The abstraction-driven heuristic is one method for exploiting this choice point. Similarly, the minimum conspiracy number heuristic and the SABA heuristic are not the only ways to guide the choices of instantiated operator and applicable operator respectively. The heuristics used can always be changed, and we do not claim that the ones we provide as defaults are the best possible: no heuristic will work all the time.

The planning algorithm we present is both sound and complete if it searches the entire search space, using a technique such as iterative deepening (Korf, 1985). FLECS is sound because it only terminates when it has reached the goal statement as a result of applying operators. That is, the application of the operator sequence returned as the final plan has been entirely simulated by the time the planner terminates. Thus the preconditions of each operator will all be true at the time the operator is executed, and after all operators have been executed, the goal statement will be satisfied. Consequently, FLECS is sound.

Since no step in the algorithm prunes any of the search space, FLECS with an iteratively increasing depth bound is also complete: if there is a solution to a planning problem, FLECS will find one. To insure this property, we need only show that FLECS can consider all possible operators that may achieve a goal as well as all orderings of interacting applicable operators. FLECS does so by maintaining backtracking points at the choice of operator (step 6c) and at both points at which the operator ordering could be affected: the choice of applicable operator itself (step 7) and the choice of whether to subgoal or apply (step 5d). Selecting

"apply" commits to ordering all operators that are not currently applicable after at least one of the currently applicable operators. Note that completeness is achieved even without maintaining the choice of goals to subgoal on as a backtrack point (step 6), since regardless of the order in which the operators are chosen, they are applied according to their possible interactions (i.e., similarly to resolving negative threats). Thus FLECS's search space is significantly reduced from that of PRODIGY4.0, while still preserving completeness. (See Appendix A for formal proofs of FLECS's soundness and completeness.)

## 5. Empirical Analysis of Heuristics to Control the Commitment Strategy

As we have seen, FLECS introduces the notion of a flexible choice point between delayed and eager operator-ordering commitments. To appreciate the need for this flexibility, consider the two extreme heuristics: always eagerly subgoaling (delaying commitment) and always eagerly applying (eager commitment). The former heuristic chooses to subgoal as long as there is at least one active pending goal (*S*ubgoal *A*lways *B*efore *A*pplying or SABA); the latter chooses to apply as long as there are any active applicable operators (*S*ubgoal *A*fter e*V*ery *T*ry to *A*pply or SAVTA). In this section we show empirical results that demonstrate that both of these extremes can lead to highly sub-optimal search in particular domains. Indeed, we believe that no single domain-independent search heuristic can perform well in all domains (Stone et al., 1994). It is for this reason that we have equipped FLECS with the ability to use either extreme domain-independent heuristic or any more moderate heuristic "in between" the two: during every iteration through our algorithm, there is an opportunity to change from eagerly subgoaling to eagerly applying or vice versa. One could define different heuristics to guide this choice, or one could leave the choice up to the user interactively.

This flexibility in search method provides our algorithm with the ability to search sensibly in a wide variety of domains. Any algorithm that is not so flexible is susceptible to coming across domains which it cannot handle efficiently (Barrett & Weld, 1994; Veloso & Blythe, 1994; Kambhampati, 1994). FLECS's flexibility makes it possible to study which heuristics work best in which situations. In addition, this flexible choice is a perfect learning opportunity. Since no single search method will solve all planning problems, we will use learning techniques to help us determine from experience which search strategies to try.

To illustrate the need for different search strategies, we provide one real world situation in which eagerly subgoaling leads directly to the optimal solution, one in which eagerly applying does so, and one in which an intermediate policy is best. These examples are not intended to be an exhaustive demonstration of FLECS's capabilities. Rather, our examples are intended to illustrate the need to consider problems other than traditional goal ordering problems and to motivate the potential impact of FLECS.

### 5.1 Eagerly Subgoaling Can Be Better

First, consider the class of tasks in which the following is true: all operators are initially executable, but they must be performed in a specific order because each operator deletes the preconditions of the operators that were supposed to be executed earlier. For instance, suppose that there is a single paint brush and several objects which need to be painted different colors. The paint brush can be washed fairly well, but it never comes completely clean. For this reason, if we ever use a lighter paint after a darker paint, some of the darker paint will show up on the painted object and our whole project will be ruined. Perhaps the shade of red is darker than the shade of green. Then to paint a chair with a red seat and green legs, we had better paint the legs first.

Consider a range of colors ordered from light to dark: white, yellow, green, ..., and black. Initially, we could paint an object any color. However, if we start by painting something black, then no other paint can be used. In order to represent this situation to a planner, we created a domain with the operators shown in Table 3.

| **Operator**: | **paint-white** <obj> | **paint-yellow** <obj> | ··· | **paint-black** <obj> |
|---:|---|---|---|---|
| **preconds**: | (usable white) | (usable yellow) | ··· | (usable black) |
| **adds**: | (white <obj>) | (yellow <obj>) | ··· | (black <obj>) |
| **deletes**: | | (usable white) | ···<br>···<br>⋮ | (usable white)<br>(usable yellow)<br>⋮<br>(usable brown) |

Table 3: Example domain for which delayed step-ordering commitment results in efficient planning.

Assume that all the colors are usable in the initial state. Since painting an object a certain color deletes the precondition of painting an object a lighter color, and since this precondition cannot be re-achieved (no operator adds the predicate "usable"), the colors must be used in a specific order.

This painting domain is a real-world interpretation of the artificial domain $D^m S^1$ introduced in (Barrett & Weld, 1994). The operators in $D^m S^1$ look like:

$$\begin{aligned} \text{Operator:} \quad & A_i \\ \text{preconds:} \quad & \{I_i\} \\ \text{adds:} \quad & \{G_i\} \\ \text{deletes:} \quad & \{I_j \mid j < i\} \end{aligned}$$

Since each operator deletes the preconditions of all operators numerically before it, these operators can only be applied in increasing numerical order. Thus, $A_1$ corresponds to the operator paint-white, $A_2$ corresponds to paint-yellow, etc. We used this domain for our experiments, all of which were run on a SPARC station. We generated random problems having one to fifteen goals: ten problems with each number of goals. We used these same 150 problems to test both of the extreme heuristics. To get our data points, we averaged the results for the ten problems with the same number of goals. All of the raw data is contained in the online appendix. We graph the average time that FLECS took to solve the problems versus the number of goals.

As shown in (Stone et al., 1994),[9] eagerly applying leads to exponential behavior (as a function of the number of goals) in this domain, while eagerly subgoaling, when using

9. We began the study of our new planning algorithm — now named FLECS— on PRODIGY4.0. We consider the version of PRODIGY used in (Stone et al., 1994) to be a preliminary implementation of FLECS.

an operator choice heuristic from the same study, leads to approximately linear behavior and no backtracking. The problem with eagerly applying is that, for example, if goal $G_7$ is solved before $G_4$, then FLECS will immediately apply $A_7$ and have to backtrack when it unsuccessfully tries to apply $A_4$. Eagerly subgoaling allows FLECS to build up the set of operators that it will need to apply and then order them appropriately by selecting an application order that avoids conflicts or threats. Figure 11 shows a graphic comparison of the two different behaviors.

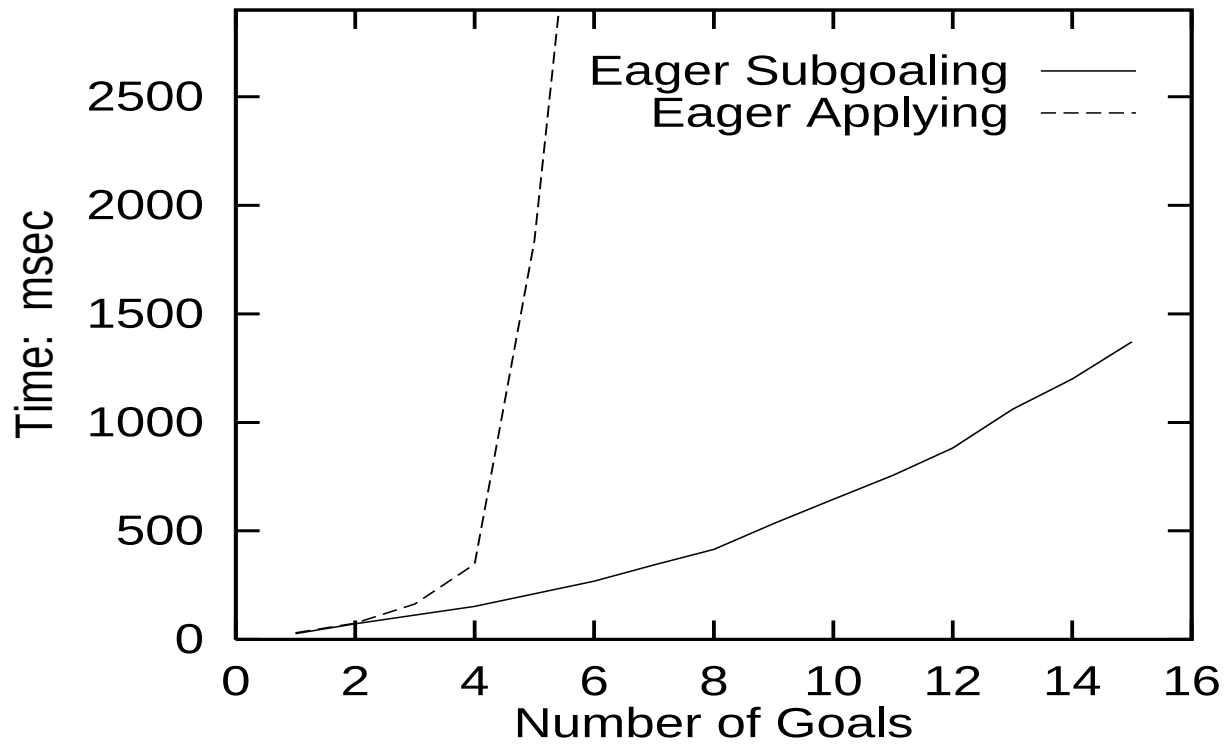

Figure 11: FLECS's performance with different heuristics in domains $D^m S^1$. Eager subgoaling and applying correspond to delayed commitments and eager commitments respectively.

### 5.2 Eagerly Applying Can Be Better

Next, consider the class of tasks in which the following is true: several operators could be used to achieve any goal, but each operator can only be used once. To use a similar example, suppose we are trying to paint different parts of a single object different colors. However, now suppose that we are using multiple brushes that never come clean: once we use a brush for one color, we can never safely use it again. For instance, if we painted the green parts using brush1, we would need to use brush2 (or any brush besides brush1) to paint the red parts. Table 4 represents the operators in this new domain.

| **Operator**: | **paint-with-brush1** <parts> <color> | ... | **paint-with-brush8** <parts> <color> |
|---|---|---|---|
| **preconds**: | (unused brush1) | ... | (unused brush8) |
| **adds**: | (painted <parts> <color>) | ... | (painted <parts> <color>) |
| **deletes**: | (unused brush1) | ... | (unused brush8) |

Table 4: Example domain for which eager step-ordering commitment and use of the state results in efficient planning.

Note that each operator can be used for any color, but since it deletes its own precondition, it can only be used once. We capture the essential features of this domain in an artificial domain called $D^1$-use-once. The operators in $D^1$-use-once look like:

| | |
|---:|:---|
| Operator: | $A_i$ |
| preconds: | $\{I_i\}$ |
| adds: | $\{< g >\}$ |
| deletes: | $\{I_i\}$ |

Any operator can achieve any goal, but since each operator deletes its own precondition, it can only be used once. Each operator corresponds to painting with a different brush.

In this domain, it is better to eagerly apply than it is to eagerly subgoal. Eagerly subgoaling causes FLECS to select the same operator to achieve all of its goals. With a deterministic method for selecting operators (such as minimum conspiracy number with order of appearance in the domain specification as a tie-breaker), it selects operator $A_1$ to achieve two different goals. However, since it could only apply $A_1$ once, it would need to backtrack to select a different operator for one of the goals. As shown in Figure 12, eagerly applying outperforms eagerly subgoaling in this case. We generated these results in the same way as the results in the previous subsection.

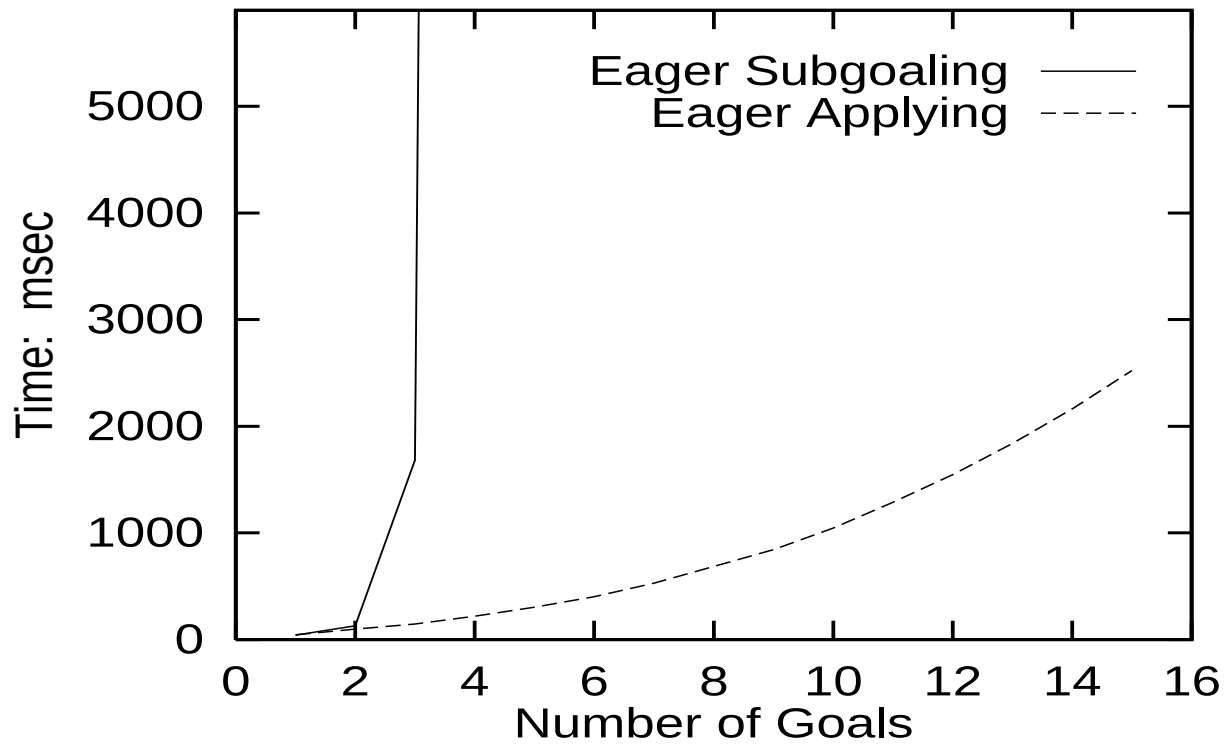

Figure 12: FLECS's performance with different heuristics in domains $D^1$-use-once.

### 5.3 An Intermediate Heuristic

Were it always possible to find good solutions either by always eagerly subgoaling, as in the first example, or by always eagerly applying, as in the second, there would be no need to include the variable *toggle* in FLECS: we could simply have an eager-subgoal mode and an eager-apply mode. However, there are cases when neither of the above alternatives suffices. Instead, we need to eagerly subgoal during some portions of the search and eagerly apply during others. One heuristic for changing the commitment strategy is the abstraction-driven method described in Section 4.2. Here we present a domain which can use a form of this heuristic.

This time consider the class of tasks in which the following is true: top-level goals take at least three operators to achieve, one of which is irreversible, can only be executed a limited number of times, and restricts the bindings of the other operators. One representative of this class is the one-way rocket domain introduced in (Veloso & Carbonell, 1993). For the sake of consistency, however, we will present a representative of this class of domains in the painting context. Suppose that we are painting walls with rollers. To paint a wall we need to first "ready" the wall, which for the purpose of this example means to decide that the wall needs to be painted and to designate a color and roller to paint the wall. Next we must fill the selected roller with the appropriately colored paint. Then we can paint the wall. Unfortunately, our limited supply of rollers can never become clean after they have been filled with paint, but they must be clean when they are selected to paint a wall. For this reason, we must ready all the walls that we want to paint with the same roller *before* we can fill the roller with paint. For the reader familiar with the one-way rocket domain, the "fill-roller" operator here is analogous to the "move-rocket" operator in that domain: it can only be executed once due to a limited supply of fuel, and it must be executed *after* it has been fully loaded. Table 5 shows a possible set of operators in this painting domain.

| **Operator**: | **designate-roller**<br><wall> <roller> <color> | **fill-roller**<br><roller> <color> | **paint-wall**<br><wall> <roller> <color> |
|---|---|---|---|
| **preconds**: | (clean <roller>)<br>(needs-painting <wall>) | (clean <roller>)<br>(chosen<br><roller> <color>) | (ready<br><wall> <roller> <color>)<br>(filled-with-paint<br><roller> <color>) |
| **adds**: | (ready<br><wall> <roller> <color>)<br>(chosen <roller> <color>) | (filled-with-paint<br><roller> <color>) | (painted <wall> <color>) |
| **deletes**: | | (clean <roller>) | (ready<br><wall> <roller> <color>)<br>(needs-painting <wall>) |

Table 5: Example domain for which the flexibility of commitments results in efficient planning.

When given this domain representation, FLECS has a difficult time with some apparently simple problems if it uses the same search strategy throughout its entire search. For example, consider the problem with five walls and two rollers (equivalent to a problem in the one-way rocket domain with five objects and two destinations):

| Initial State | Goal Statement | An Optimal Solution |
|---|---|---|
| (needs-painting wallA) | (painted wallA red) | <Designate-Roller wallA roller1 red> |
| (needs-painting wallB) | (painted wallB red) | <Designate-Roller wallB roller1 red> |
| (needs-painting wallC) | (painted wallC red) | <Designate-Roller wallC roller1 red> |
| (needs-painting wallD) | (painted wallD green) | <Fill-Roller roller1 red> |
| (needs-painting wallE) | (painted wallE green) | <Paint-Wall wallA roller1 red> |
| (clean roller1) | | <Paint-Wall wallB roller1 red> |
| (clean roller2) | | <Paint-Wall wallC roller1 red> |
| | | <Designate-Roller wallD roller2 green> |
| | | <Designate-Roller wallE roller2 green> |
| | | <Fill-Roller roller2 green> |
| | | <Paint-Wall wallD roller2 green> |
| | | <Paint-Wall wallE roller2 green> |

flecs does not directly find this solution when always eagerly subgoaling or when always eagerly applying. To search efficiently, it must subgoal until it has considered all the walls that need to be painted the same color; then it must apply all applicable operators before continuing. There is no explicit information in the domain telling it to use one roller for red and one roller for green.[10] For this reason, when flecs eagerly subgoals, it initially selects the same roller to paint all the walls. It extensively backtracks before finding the correct bindings. flecs also does not realize that it should "ready" all the walls that are going to be painted the same color before filling the roller. Thus, when flecs eagerly applies operators, it tries filling a roller as soon as it has one wall "readied." Note that planning with variables would not solve this problem since the planner would still need to make binding selections before subgoaling beyond "paint-wall," hence facing the same problems.

When flecs tries to solve the above problem using either strategy described, it does not succeed in a reasonable amount of time. Since flecs is complete, it would certainly succeed eventually, but eventually can be a long time away when dealing with an NP-hard problem: neither of these commitment strategies leads to a solution to the above problem in under 500 seconds of search time. But all is not lost. By changing the value of *toggle* at the appropriate times, flecs can easily find a solution to the above problem. In fact, it can do so in just 4 seconds when *toggle* is manually changed at the appropriate times.

| | time(sec) | solution |
|---|---|---|
| eager applying | 500 | no |
| eager subgoaling | 500 | no |
| variable strategy | 4 | yes |

If flecs eagerly subgoals until it has decided to paint wallA, wallB, and wallC with roller1, then it can begin eagerly applying. Once the three walls are painted red, flecs can begin subgoaling again without any danger of preparing the other walls with the wrong roller: only roller2 is still clean. This is an example in which the change in state allows the minimum conspiracy number heuristic to select the correct instantiated operator.

The general heuristic here is that *toggle* should be set to *sub* until all walls that need to be painted the same color have been considered. Then *toggle* should be set to *app* until all the applicable operators have been applied. Then *toggle* should be set back to *sub* as the process continues. In this way, flecs will need to do very little backtracking and it can quickly reach a solution. This heuristic corresponds to using an abstraction hierarchy to deal separately with the interactions between the different colors and the different walls.

## 6. Conclusion

We have presented a planner that is intended for studying the correspondence between planning problems and the search heuristics that are most suited to those problems. flecs has the ability to eagerly subgoal, thus delaying operator-ordering commitments; eagerly apply, thus maximizing the advantages of maintaining an internal state; or to flexibly interleave these two strategies. Thus it can operate at any point in the continuum of operator-ordering heuristics – one important dimension of planning.

10. This problem is very common in planning as there is often no syntactically correct way to restrict bindings in a domain representation while maintaining the intended flexibility and generality in the domain.

In this paper, we explained the advantages and disadvantages of delayed and eager commitments. We presented the FLECS algorithm in full detail, carefully motivating the concepts and illustrating them with clear examples. We discussed different heuristics to guide FLECS in its choice points and discussed its properties. We showed examples of specific planning tasks and corresponding empirical results which support our position that a general-purpose planner must be able to use a flexible commitment strategy. Although all planning problems are solvable by complete planners, FLECS may solve some of the problems more efficiently than other planners that do not have the ability to change their commitment strategy and may fall into a worst case of their unique commitment strategy.

FLECS provides a framework to study the characteristics of different planning strategies and their mapping to planning domains and problems. FLECS represents our view that there is no domain-independent planning strategy that is uniformly efficient across different domains and problems. FLECS addresses the particular operator-ordering choice as a flexible planning decision. It allows the combination of delayed and eager operator-ordering commitments to take advantage of the benefits of explicitly using a simulated execution state and reasoning about planning constraints.

We are currently continuing our work on understanding the tradeoffs among different planning strategies along different dimensions. We plan to study the effects of eager versus delayed commitments at the point of operator instantiations. We are also investigating the effects of combining real execution into FLECS. Finally, we plan to use machine learning techniques on FLECS's choice points to gain a possibly automated understanding of the mapping between efficient planning methods and planning domains and problems.

## Appendix A. Proofs

We prove that FLECS is sound and that with iterative deepening it is complete. Consider the FLECS algorithm as presented in Table 2. A *planning problem* is determined by the initial state, the goal statement, and the set of operators available in the domain. A *plan* is a (totally-ordered) sequence of instantiated operators. The *returned plan* generated by FLECS for a planning problem is the sequence of applied operators upon termination. A *solution* to a planning problem is a plan whose operators can be applied to the problem's initial state so as to reach a state that satisfies the Goal Statement. A *justified solution* is a solution such that no subsequence of operators in the solution is also a solution. FLECS *terminates successfully* when the termination condition is met (step 2).

**Theorem 1.** FLECS *is sound.*

We show that the FLECS algorithm is sound; that is, if the algorithm terminates successfully, then the returned plan is indeed a solution to the given planning problem.

Assume that FLECS terminates successfully and that $S = O_1, O_2, ...O_n$ is the returned plan. FLECS applies an operator only when the preconditions of the operator are satisfied in the Current State $\mathcal{C}$ (step 7). Hence, by construction, after operators $O_1, O_2, \ldots O_k$ for any $k < n$ have been applied, the preconditions of operator $O_{k+1}$ are satisfied in $\mathcal{C}$. At the point of termination, the Current State $\mathcal{C}$ satisfies the Goal Statement (step 2). But $\mathcal{C}$ was reached from the initial state by applying the operators of $S$. Therefore $S$ is a solution. QED.

**Theorem 2.** FLECS *with iterative deepening is complete.*

Recall that completeness, informally, means that if there is a solution to a particular problem, then the algorithm will find it. We show that FLECS's search space is complete and that FLECS's search algorithm is complete as long as it explores all branches of the search space, for example using iterative deepening (Korf, 1985).[11] Iterative deepening involves searching with a bound on the number of search steps that may be performed before a particular search path is suspended from further expansion; if no solution is found for a particular depth bound, the search is repeated with a larger depth bound.

For a planning problem, assume that $S = O_1, O_2, ...O_n$ is a justified solution. We will show that if FLECS searches with iterative deepening, it will find a solution.

The FLECS algorithm has four choice points. Three of these choice points are backtrack points: the choice between subgoaling and applying (step 5d), the choice of which operator to use to achieve a goal (step 6c), and the choice of which applicable operator to apply (step 7). One choice point is not a backtrack point: the choice of goal on which to subgoal (step 6).

To prove completeness, we must show that at each backtrack point, there is some possible choice that will lead FLECS towards finding the plan $S$, no matter what choices FLECS makes at the non-backtrack choice point. Then if FLECS explores all branches of the search space by searching with iterative deepening, it must eventually find $S$ unless it finds some other solution (of length $\leq n$) first.

The proof involves constructing *oracles* that tell FLECS which choices to make at the backtrack points so as to find $S$. Then no matter what choices it makes at the other choice point, it finds solution plan $S$.

Consider the point in the search at which operators $O_1, O_2, O_3, \ldots, O_k$ for some $k$ (and no others) have already been applied. Then let there be oracles at the backtrack points which operate as follows.

At the choice of subgoaling or applying (step 5d), the first oracle makes FLECS choose to apply if and only if $O_{k+1}$ is applicable (i.e., is in $\mathcal{A}$); otherwise it makes FLECS subgoal. If FLECS chooses to apply ($O_{k+1} \in \mathcal{A}$), then it reaches another choice point, namely the choice of operator to apply (step 7). Another oracle makes FLECS select precisely the step $O_{k+1}$.

If FLECS chooses to subgoal ($O_{k+1} \notin \mathcal{A}$), then let FLECS choose any goal $P$ from the set of pending goals $\mathcal{P}$ (step 6). Since step 6 is not a backtrack point, we cannot have an oracle determine the choice at this point. Instead we have to show that, independently from the choice made at this point, FLECS will still find the solution $S$. It can find this solution as a consequence of the construction of the next oracle that controls the final choice point (below). That oracle guarantees that any $P$ selected must either be a member of the goal statement or a precondition of some operator of $S$.

The final choice point is the selection of an operator to achieve $P$ (step 6c). The third oracle makes FLECS choose an operator of $S$ to achieve $P$. Since $S$ is a solution to the planning problem and since $P$ is either a member of the Goal Statement or a precondition of some operator of $S$, there must be some operator of $S$ that achieves $P$. If there is more than one such operator, any one can be chosen. Since only operators from $S$ are selected,

11. As opposed to breadth first search, iterative deepening does not harm efficiency. It combines the efficiency of searching depth first with the completeness of searching breadth first.

the condition that all pending goals are from the Goal Statement or are preconditions of operators of $S$ is maintained.

These three oracles will lead FLECS to the justified solution $S$. Since $S$ is justified, every operator of $S$ is necessary to achieve either some goal in the goal statement or some precondition of another operator. Consequently, since the third oracle only chooses operators of $S$, every such operator will eventually be chosen and then applied as prescribed by the first two oracles. Once every operator of $S$ has been applied, the termination condition will be met (since $S$ is a solution) and FLECS will terminate successfully. QED.

## Acknowledgements

We would like to recognize in particular the contributions of Jim Blythe and Eugene Fink to our research. Jim Blythe is highly responsible for the current implementation of PRODIGY4.0 upon which FLECS is based. Eugene Fink helped with the formalization of our algorithms and proofs. We thank Eugene Fink, Karen Haigh, Gary Pelton, Alicia Pérez, Xuemei Wang, and the anonymous reviewers for their comments on this article.

This research is sponsored by the Wright Laboratory, Aeronautical Systems Center, Air Force Materiel Command, USAF, and the Advanced Research Projects Agency (ARPA) under grant number F33615-93-1-1330. The views and conclusions contained in this document are those of the authors and should not be interpreted as necessarily representing the official policies or endorsements, either expressed or implied, of Wright Laboratory or the U. S. Government.